\documentclass[letterpaper, 10 pt, conference]{ieeeconf}
\IEEEoverridecommandlockouts
\let\labelindent\relax
\usepackage{graphicx} \graphicspath{ {figures/} }
\usepackage{amsmath,amssymb,mathabx}
\usepackage{acronym}
\usepackage{enumitem}
\usepackage{booktabs}
\usepackage{hyperref}
\usepackage{balance}
\usepackage{xspace,setspace}
\usepackage[skip=3pt,font=small]{subcaption}
\usepackage[skip=3pt,font=small]{caption}
\usepackage[dvipsnames]{xcolor}
\usepackage[capitalise]{cleveref}
\usepackage{tabularx,colortbl,multirow,array,makecell}
\usepackage{cite}

\makeatletter
\DeclareRobustCommand\onedot{\futurelet\@let@token\@onedot}
\def\@onedot{\ifx\@let@token.\else.\null\fi\xspace}
\def\eg{\emph{e.g}\onedot} 
\def\ie{\emph{i.e}\onedot}

\def\etal{\emph{et al}\onedot}
\makeatother

\crefname{algocf}{alg.}{algs.}
\Crefname{algocf}{Algorithm}{Algorithms}

\makeatletter
\def\BState{\State\hskip-\ALG@thistlm}
\makeatother

\makeatletter
\renewcommand{\paragraph}{%
  \@startsection{paragraph}{4}%
  {\z@}{0ex \@plus 0ex \@minus 0ex}{-1em}%
  {\hskip\parindent\normalfont\normalsize\bfseries}%
}
\makeatother

\crefname{algocf}{alg.}{algs.}
\Crefname{algocf}{Algorithm}{Algorithms}

\definecolor{gblue}{HTML}{4285F4}
\definecolor{gred}{HTML}{DB4437}

\acrodef{lfd}[LfD]{Learning from Demonstration}
\acrodef{rl}[RL]{Reinforcement Learning}
\acrodef{il}[IL]{Imitation Learning}
\acrodef{drl}[DRL]{Deep Reinforcement Learning}
\acrodef{vln}[VLN]{Vision-Language Navigation}
\acrodef{hai}[HAI]{Human-AI Interaction}
\acrodef{vqa}[VQA]{Visual Question Answering}
\acrodef{ppo}[PPO]{Proximal Policy Optimization}
\acrodef{mdp}[MDP]{Markov Decision Process}
\acrodef{cnn}[CNN]{Convolutional Neural Network}
\acrodef{sr}[SR]{Success Rate}
\acrodef{spl}[SPL]{Success weighted by Path Length}
\acrodef{gesthor}[Ges-THOR]{Gesture-based THOR}
\acrodef{gru}[GRU]{Gated Recurrent Unit}
\usepackage{etoolbox,siunitx}

\title{\LARGE \bf Communicative Learning with Natural Gestures\\for Embodied Navigation Agents with Human-in-the-Scene}

\author{Qi Wu$^1$\quad{}Cheng-Ju Wu$^1$\quad{}Yixin Zhu$^1$\quad{}Jungseock Joo$^{1\star}$\quad{}
\thanks{$^\star$Corresponding author. Email: \tt{jjoo@comm.ucla.edu}}
\thanks{$^1$UCLA. Emails: \tt\{qi.wu,jimmy.wu,yixin.zhu\}@ucla.edu}%
\thanks{To appear in 2021 IEEE/RSJ International Conference on Intelligent Robots and Systems (IROS). This work was supported by NSF NRI \#1925360.}%
}
\robustify\bfseries
\begin{document}

\maketitle
\thispagestyle{empty}
\pagestyle{empty}

\begin{abstract}
Human-robot collaboration is an essential research topic in artificial intelligence (AI), enabling researchers to devise cognitive AI systems and affords an intuitive means for users to interact with the robot. Of note, communication plays a central role. To date, prior studies in embodied agent navigation have only demonstrated that human languages facilitate communication by instructions in natural languages. Nevertheless, a plethora of other forms of communication is left unexplored. In fact, human communication originated in gestures and oftentimes is delivered through multimodal cues, \eg, ``go there'' with a pointing gesture. To bridge the gap and fill in the missing dimension of communication in embodied agent navigation, we propose investigating the effects of using gestures as the communicative interface instead of verbal cues. Specifically, we develop a VR-based 3D simulation environment, named \ac{gesthor}, based on AI2-THOR platform. In this virtual environment, a human player is placed in the same virtual scene and shepherds the artificial agent using only gestures. The agent is tasked to solve the navigation problem guided by natural gestures with unknown semantics; we do not use any predefined gestures due to the diversity and versatile nature of human gestures. We argue that learning the semantics of natural gestures is mutually beneficial to learning the navigation task---\textit{learn to communicate and communicate to learn}. In a series of experiments, we demonstrate that human gesture cues, even without predefined semantics, improve the object-goal navigation for an embodied agent, outperforming various state-of-the-art methods.
\end{abstract}

\setstretch{0.97}

\section{Introduction}

Human-human communication takes place in various forms, of which gestures play a crucial role~\cite{joo2017red}. Gestures include movements of body, head, or hands and can facilitate the understanding of the speech or serve as emblems to deliver messages in place of speech~\cite{joo2019automated,kang2020understanding}. They can significantly improve the communication efficacy for information conveyance~\cite{bremner2016iconic}.

Similarly, human-robot communication can also occur using multimodal cues~\cite{zhu2020dark}. Although robots and autonomous systems are designed to collaborate with humans who supervise, instruct, or evaluate the system to perform specific tasks, most existing communication interfaces assume that humans communicate to an artificial agent only using natural language, either verbally or through text. In stark contrast, the origin of human communications is primarily rooted in nonverbal forms~\cite{tomasello2010origins}, \eg, gestures. Therefore, providing assistive or collaborative AI systems with nonverbal means of communication would open up new research venues to investigate the efficacy of alternative communication forms. Unlike natural languages, which suffer from intermittent conveyance and need continuous attention, nonverbal cues like gestures are immediate and intuitive, hence are less vulnerable to interruptions. In particular, when the environment is noisy, or the agent is listening to someone else, a user might refer to a location using a deictic gesture, \eg, pointing with a finger, instead of describing it with a long sentence.

\begin{figure}[t!]
    \centering
    \includegraphics[width=0.9\linewidth]{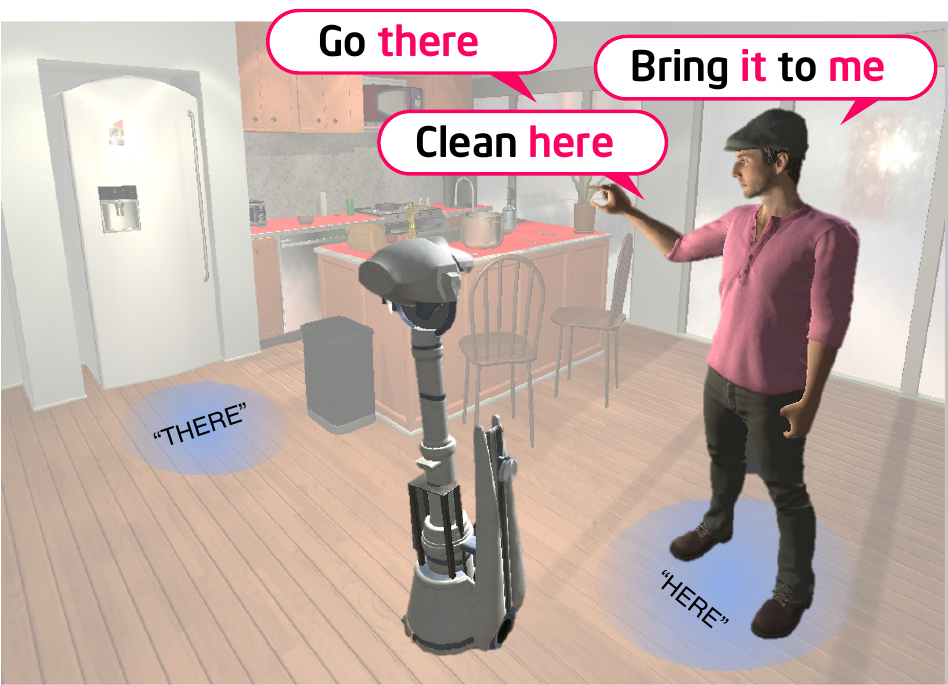}
    \caption{\textbf{Natural gestures can succinctly deliver complex semantic meaning in a physical space.} A human user can instruct a robot or a virtual agent to complete a navigation task by simply referring to a target location or object using gestures. The agent should infer the user intent from the gestures.  }
    \label{introduction}
    \label{fig:intro}
\end{figure}

To illustrate the significance of communications using gestures, let us take \cref{fig:intro} as an example. A human intends to instruct the robot to navigate to the target location or object in the scene. 
Previous works in embodied visual navigation with language-based human interactions may require a lengthy text message, such as ``go to the second brown chair next to the big white table in the living room''. In contrast, gestures allow to express the same message in a much simpler and more natural way, \eg ``go there,'' ''clean here,'' or ''bring it to me.'' Such multimodal messages can only be correctly interpreted in a given physical space where a human and an agent are situated together. The meaning of a human message must be inferred from the joint understanding of the given scene by the agent who also understands the semantics of human natural gestures~\cite{steen2013multimodal,steen2018toward}.


Inspired by the above crucial observation, we intend to bring in nonverbal communication cues~\cite{joo2017red,fan2021learning,joo2019towards} into the embodied agent navigation task---the most straightforward task of an embodied AI that interacts with the environments and other agents. Despite the progress reported for the embodied agent on the \ac{vln} task~\cite{chen2011learning,chaplot2017gated,gupta2017cognitive,das2018embodied,anderson2018vision,wang2019reinforced,chen2019touchdown,fried2018speaker,zhu2020vision-language}, we contemplate on prior arts and quest for the following questions: Instead of using natural language, can we replace the language grounding by gestures in a similar setting? Can we improve the performance of navigation with gestures incorporated? Can the learning agent acquire the underlying semantics of gestures, even when they are not predefined?

Specifically, we aim to use gestures to communicate with an embodied agent to navigate in a virtual environment. To provide gesture-based instructions for a navigation task, the agent needs a photorealistic simulation environment, and a human player needs to be situated in the same scene to have \textit{joint attention}~\cite{tomasello1986joint}. To support such a co-existing environment, we build our virtual environment \ac{gesthor} with Oculus, Kinect, and Leap Motion, based on the existing framework AI2-THOR~\cite{kolve2017ai2thor}.

Although human gestures have been used as a communicative interface between humans and robots in robotics~\cite{hasanuzzaman2004real,nickel2007visual,ertuugrul2013gesture,nuzzi2019deep,chang2019improved}, prior literature typically predefines the vocabulary of admissible gestures and their definite meanings (\eg, ``ok'' sign means an approval). In contrast, human gestures are diverse; their meanings are also non-rigid and context-dependent~\cite{jiang2021individual}.
One needs to develop a flexible system to address the variability and versatility of nearly-unlimited naturalistic human gestures without a predefined set of recognizable gestures. 
Without defining any gestures and their meanings ourselves, we have collected demonstrations from a group of volunteers who have diverse gesture preferences for the same message. 

In our proposed framework, an agent should therefore solve two tasks: multimodal target inference and navigation. Inferring the meanings of human gestures and finding a path to the target location are two major goals of the agent, which mutually help each other, \ie \textbf{\textit{learn to communicate and communicate to learn.} }
Experiments reveal that our model incorporating gestures outperforms a baseline model only with vision for navigation, as well as models on similar environments and tasks using different methods~\cite{yang2018visual}.

This paper makes four contributions: (i) By introducing human gestures as the new communicative interface for embodied AI learning and (ii) developing a simulation framework, \ac{gesthor}, that supports multimodal interactions with human users, (iii) we demonstrate that the embodied agent's navigation performance significantly improves after incorporating human gestures. (iv) We further demonstrate that the agent can learn the underlying meanings and intents of human gestures without predefining the associations.

\section{Related Work}


\paragraph*{Language grounding}

Language grounding is crucial for both parties involved in communication to understand each other. Natural language, the most common modality for human-human and human-robot communication, can realize the grounding in various ways. For communication with robots, language can be interpreted from instructional commands to actions~\cite{matuszek2013learning,misra2018mapping}. For static images or texts, it can be either visually grounded~\cite{antol2015vqa,niu2019recursive} or text-based~\cite{weston2015towards} Q\&A. In our work, language grounding is replaced by ``gesture grounding;'' we provide gestures as the new communicative interface. The agent is tasked to learn by grounding human gestures into a series of actions and identify target objects.

\paragraph*{Vision-language navigation}
Image captioning with large datasets~\cite{vinyals2016show} and \ac{vqa}~\cite{antol2015vqa,marino2019okvqa} has made significant progress in vision and language understanding, which enables visually-grounded language navigation agent to be trained. Many tasks following the \ac{vln} framework~\cite{chen2011learning,chaplot2017gated,das2018neural,das2018embodied,anderson2018vision,wang2019reinforced,chen2019touchdown,fried2018speaker,thomason2020vision,hao2020towards,zhu2020vision-language,shridhar2020alfworld} have been addressed and solved using end-to-end learning models, either in 2D world~\cite{yu2018interactive,chevalier2018babyai,cao2020babyai++}, 3D world~\cite{Hermann2017grounded,chaplot2017gated}, or even photorealistic environments~\cite{anderson2018vision,chen2019touchdown,fried2018speaker,xie2019vrgym,zhu2020vision-language}. Some works have also explored the acoustic cue in navigation, but these are not mainly concerned with speech~\cite{gan2020look,chen2020soundspaces}. 
Our work is built on the existing \ac{vln} framework but extended by incorporating gestures as a new modality for communications.

\paragraph*{Simulated environments}

To help the research in embodied AI learning, various simulated environments have spurred for the community's benefit. Those 3D environments are created from either synthetic scenes~\cite{brodeur2017home,wu2018building,kolve2017ai2thor,xie2019vrgym,shridhar2020alfred,puig2018virtualhome} or real photographs~\cite{savva2017minos,xia2018gibson,savva2019habitat,chang2017matterport3d}; some of them use game engines to enable physical interactions~\cite{kempka2016vizdoom,beattie2016deepmind,kolve2017ai2thor,yan2019chalet,xie2019vrgym}. In this paper, we choose AI2-THOR, which uses Unity as the physics engine, and build the environment on it. Exiting works using AI2-THOR for visual navigation tasks~\cite{zhu2017target,yang2018visual} require either the target visualization or its context. In this paper, we propose a gesture-based method to eliminate the need for acquiring additional target information.

\paragraph*{RL for navigation}

Instead of using traditional path-planning approaches~\cite{kavraki1996probabilistic} to compute a route to the goal location, the embodied AI community has recently focused more on end-to-end learning for training navigation policies, especially with \ac{rl}. Compared with other machine learning methods, such as supervised learning~\cite{mousavian2019visual}, \ac{rl} benefits from simple reward definitions and easy implementations. As a result, \ac{rl} becomes the core of the learning framework~\cite{savva2019habitat,xia2018gibson,anderson2018vision,zhu2017target,gupta2017cognitive,yang2018visual}. In this paper, we choose \ac{ppo}~\cite{schulman2017proximal} as the \ac{rl} model.

\paragraph*{Human AI interaction}

\ac{hai} has been intensively investigated in AI, Human-Computer Interaction, and robotics~\cite{goodrich2008human,amershi2019guidelines,kiesler2008anthropomorphic}. For the embodied navigation agents, the sprout of simulated environments makes users communicate with the agent interactively. Most existing frameworks achieve this goal using dialogues~\cite{chai2018language,das2018embodied,gordon2018iqa,yu2018interactive,nguyen2019help,thomason2020vision}. However, as discussed, natural language is not the only cue for multimodal communication, and current collaborative frameworks have not yet fully explored a rich spectrum of communicative interfaces for embodied agent navigation. In this paper, we propose gestures as the communicative interface between human users and the artificial agent.

Meanwhile, there is a large body of work on human gestures as a communicative device either to humans or robots~\cite{chen2015hand,gao2017static}. Most of these approaches are based on a predefined gesture set with fixed meanings or focus on gesture type classification~\cite{desemdt2016skeleton}, pose estimation~\cite{ge2018hand,baek2019pushing}, or both~\cite{yang2020collaborative,garcia2018first}. In contrast, we let users use any natural gestures and demonstrate the agent can directly learn the semantics and underlying intents of these gestures.

\section{\texorpdfstring{\ac{gesthor}}{}: A Simulation Framework for Human-Agent Interaction via Gestures}

We build an interactive learning framework in Unity based on the iTHOR environment from AI2-THOR for the gesture-boosted embodied AI research, namely \ac{gesthor}---Gesture-based iTHOR environment. 

\subsection{Simulation and Learning Framework}

There are many existing physics-based simulation frameworks for photorealistic indoor navigation tasks~\cite{xia2018gibson,xie2019vrgym,savva2019habitat,kolve2017ai2thor,savva2017minos,yan2019chalet}. We choose AI2-THOR specifically to build our learning environment because it provides a diversity of rooms and interactive features. It has been widely used for different visual navigation tasks~\cite{zhu2017target,yang2018visual}. In addition, the game engine Unity provides the ability to deploy across platforms and integrate third-party resources, compatible with the sensory devices we use for this learning environment. We also use AllenAct~\cite{luca2020allenact} as the codebase for our modular framework.

\begin{figure}[t!]
    \centering
    \vspace{-10pt}
    \includegraphics[width=\linewidth]{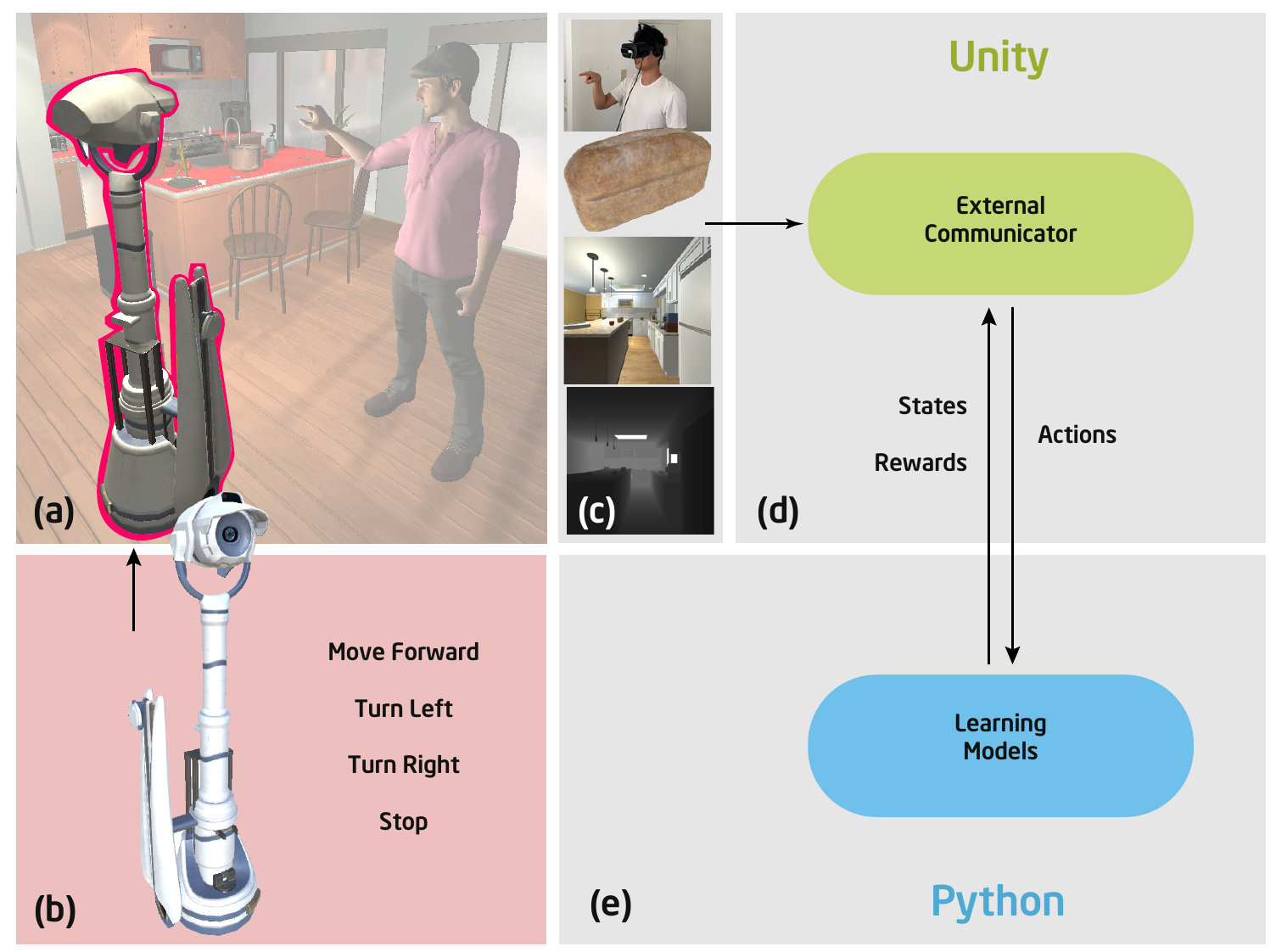}
    \caption{\textbf{Overview of the learning framework.} (a) Scenes and the (b) learning agent are built in Unity. The agent can perform four actions: move forward, turn left, and turn right, and stop. (c) It receives several sensory inputs, including RGBD images, target labels, and human gestures. Unity contains (d) an external communicator that can communicate with the (e) learning model in PyTorch. The learning model receives states and rewards from the communicator and sends back chosen actions.}
    \label{environment}
\end{figure}


\subsection{Human Gesture Sensing}

The following setup immerses human players into the virtual environment while allowing the system to capture human gestures:

\paragraph*{Devices}

We use Oculus Rift, Kinect Sensor v2, and Leap Motion Controller (hereinafter referred to as Oculus, Kinect, and Leap Motion) together for gesture sensing via pose estimation. Oculus gives the player the first-person view in the virtual environment; hence the player sees the virtual scene and knows where the target object is. Kinect is used to track overall body movements. However, Kinect is incapable of capturing fine in-hand motions. Leap Motion is brought in to detect hand movements.

\begin{figure}[t!]
    \centering
    \vspace{-10pt}
    \includegraphics[width=\linewidth]{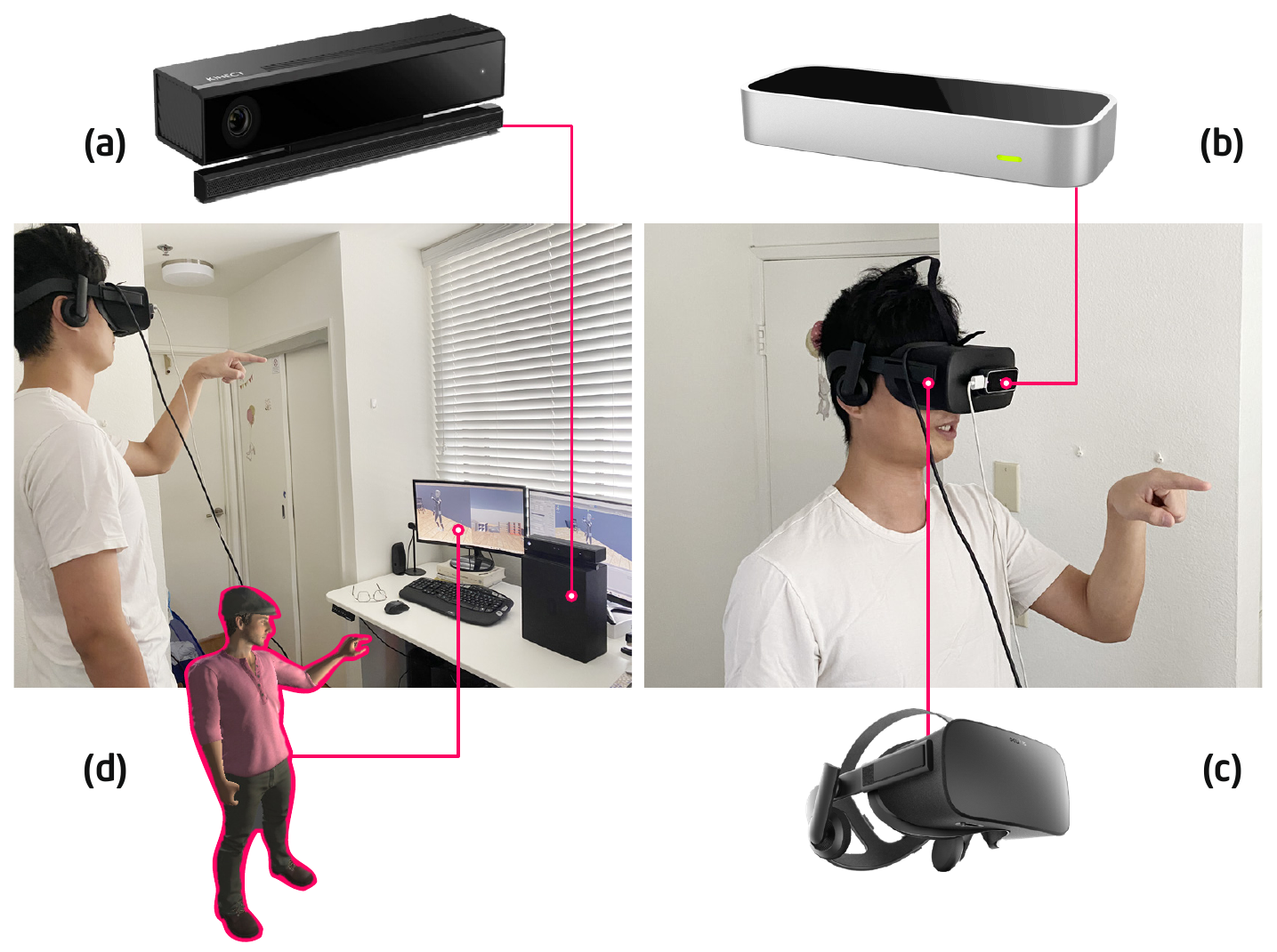}
    \caption{\textbf{Device arrangement.} The human player wears an (c) Oculus headset with (b) Leap Motion. (a) Kinect is placed 1.5m from the human player and 1.5m above the ground. The screen displays (d) A humanoid model that are mirroring the body and hand movements.}
    \label{exp_setup}
\end{figure}

\paragraph*{Device Arrangement}

\cref{exp_setup} illustrates the device arrangement. During data acquisition, the human player is asked to wear the Oculus headset, face the Kinect sensor, and move hands in front of Leap Motion at a distance between 30cm and 60cm. In Unity, a humanoid character (see \cref{exp_setup}d) mirrors players' movements in real-time, including body composure and hand motions.

\begin{figure*}[t!]
    \centering
\vspace{-10pt}    
    \includegraphics[width=0.95\linewidth]{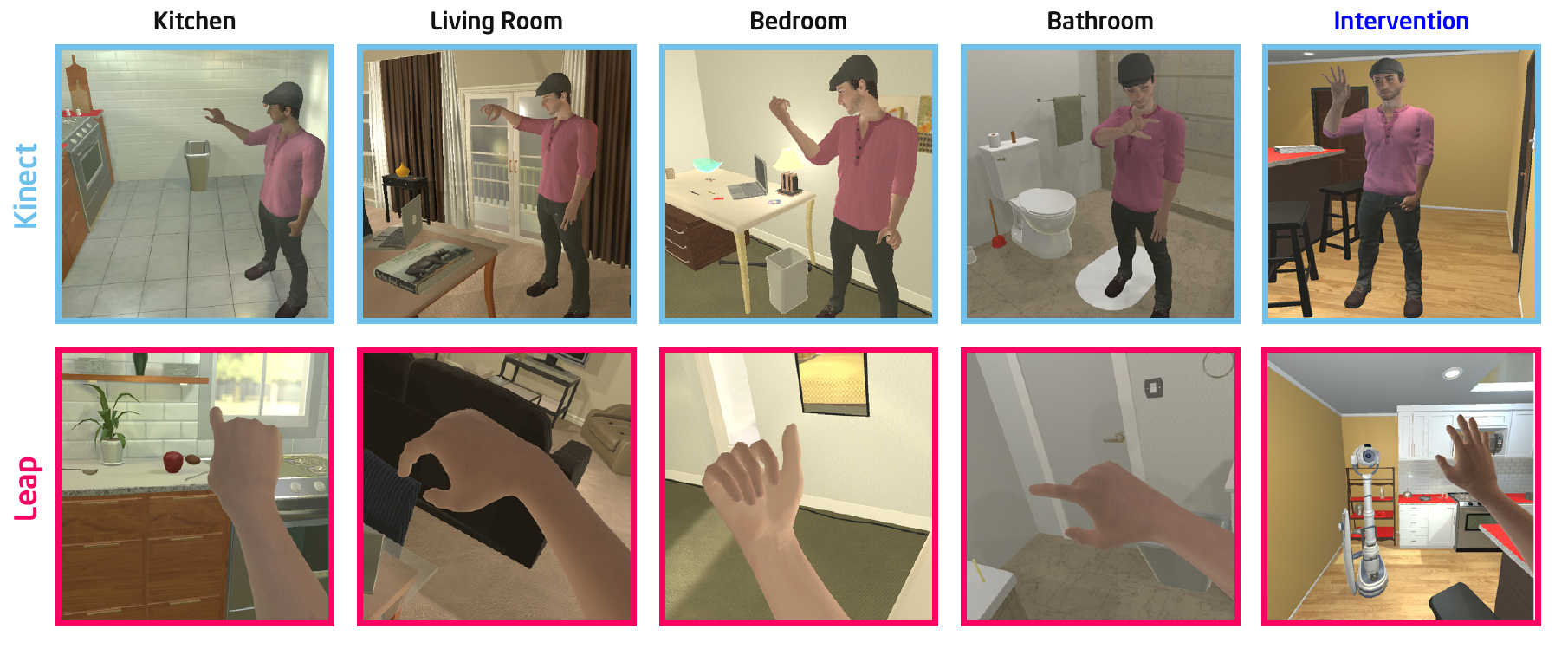}
\vspace{-10pt}       
    \caption{\textbf{Examples of referencing and intervention gestures.} The first 4 columns are referencing gestures, while the last one is an intervention gesture. Human players perform different gesture styles while pointing at various target objects in the scene. Top row shows the body movements captured by Kinect, and the bottom row shows the hand configurations recorded by Leap Motion. }
    \label{gesture_demos}
\end{figure*}

\paragraph*{Data collection}


Ideally, learning would take place in real-time, where a human player continuously observes an agent's behavior and interacts with it such that the agent can respond to the feedback immediately. Unfortunately, this is infeasible because the entire training process may take hundreds of thousands of episodes. Therefore, we opt for using pre-recorded gestures to simulate real-time interactions between the human player and the agent as closely and efficiently as possible. There are two types of instructional gestures that humans can use: one is for referencing, and the other one is for intervention. To record referencing gestures, a volunteer is given the target object in a scene and asked to communicate with the agent to guide the direction with a gesture. We do not ask participants to use any specific gestures such as pointing with a finger but encourage them to use any gestures as if they are talking to another person. 
The gesture sequence, as well as the environmental information, is recorded as one episode in the dataset. We have over 230,000 unique episodes for training and 2,500 episodes for validation and testing. For intervention gestures, the player shows gestures in a rejective manner used to warn the agent if it is moving away from the target. We recorded ten different intervention gestures. Kinect Body and Leap Hands can duplicate the player's movements and save the recorded motions as animation clips in Unity. See \cref{gesture_demos} for examples of collected gestures. 

\subsection{Sensory Modalities}

Multimodal perception is essential for artificial systems. We provide several sensory inputs in our environment to build a multimodal learning framework; see \cref{environment}. The observational space consists of the following inputs:



\paragraph*{Vision}

Unity's built-in camera component allows a 2D view of the virtual space. It is attached to the eyesight of the embodied agent at 1.5m from the ground with a 90-degree field-of-view and provides real-time RGB images in the first-person view. The resolution of the RGB images is $3\times224\times224$, and each pixel contains scaled RGB values from 0 to 1.

\paragraph*{Depth}

The depth image is extracted from the depth buffer of Unity's camera view. It has a size of $1\times224\times224$, and each pixel value is a floating-point between 0 and 1, representing the distance from the rendered viewpoint to the agent, linearly interpolated from 0m to 10m.

\paragraph*{Collision}

Unity checks for collisions dynamically in the learning environment. Every time the agent triggers a collision, it can report this event and prevent the agent from penetrating into the object meshes. Note that for our agent design, it can slide along the object surface it collides with. This ``sliding'' mechanics has been noticed by recent work~\cite{batra2020objectnav} and may hinder sim2real transition. We rectify this issue by addressing penalties in rewards for such behaviors.

\paragraph*{Gesture}


As previously mentioned, we use Oculus, Kinect, and Leap Motion to capture human gestures. Each gesture motion is saved as a sequence of vectors with 100 steps and 95 features consisting of body and hand poses. Note that for referencing gestures, we select motions from the corresponding episode. For intervention gestures, we randomly sample one from saved recordings and use it only when the agent faces away from the target. The raw gesture inputs are encoded and piped into our learning model.

\section{Learning to Navigate with Gesture}

In this section, we describe our end-to-end gesture learning model using \ac{drl}. We start by introducing the formulation of the \ac{drl} model we use, followed by the other components of the entire architecture.

\subsection{Problem Formulation}

We take the ObjectGoal task~\cite{anderson2018evaluation} as our navigation task, where the agent must navigate to an object of a specific category, as our experimental testbed. The details of the task and the agent embodiment are explained below:

\paragraph*{Agent Embodiment}

The learning agent is represented by a robot character with a capsule bound. The agent has a rigid body component attached to it so that it can detect collisions with environmental objects. It has four available actions: \textit{turn left}, \textit{turn right}, \textit{move forward}, and \textit{stop}. Each turning action results in a rotation of $15^{\circ}$, and each forward action results in a forward displacement of 0.25m.

\paragraph*{Task Definition}

The agent is initiated at a random location, and an object is selected randomly as the target; we ensure that the agent can reach the target. Note that there can be more than one instance of the target object type in the same environment. To complete the task, the agent must navigate to the target object instance with a stopping distance equal to or less than 1.5m. The agent then needs to issue a termination (\ie, \textit{stop}) action in the proximity of the goal, and the object must also be within the agent's field of view in order to succeed. An episode is terminated if the above success criteria are met or the maximum allowed time step (which is 100 in our setup) is reached. We allow the agent to issue multiple stops in an episode but measure success rates using different numbers of maximum stops (1-3). We allow an unlimited number of stops in training; the agent needs to explore and learn after issuing incorrect stops in earlier episodes.

\subsection{Policy Learning with \texorpdfstring{\ac{ppo}}{}}

We formulate our visual gesture navigation using \ac{drl}, specifically \ac{ppo}. Our learning process can be viewed as a \acf{mdp}. At each time step $t$, the agent perceives a state $s_{t}$ (\ie, a combination of the sensory inputs), receives a reward $r_{t}$ from the environment, and chooses an action $a_t$ according to the current policy $\pi$:
\begin{equation}
    a_{t} \sim \pi_{\theta}(a_{t}|s_{t}),
\end{equation}
where $\theta$ represents parameters for the function approximator of the policy $\pi$. We implement \ac{ppo} with a time horizon of 128 steps, batch size of 128 and 4 epochs for each iteration of gradient descent, and buffer size of 1280 for each policy update. We use Adam~\cite{kingma2014adam} as the optimizer with a learning rate of 0.0003 and a discount factor of 0.99.

The agent receives a positive reward of $+1$ if it completes the navigation successfully. Since we encourage the agent to reach the target object with the minimal amount of steps, the agent receives a small time penalty of $-0.001$ for each step. We further add a collision penalty of $-0.005$ for each collision detected; the collision penalty is added to mitigate the aforementioned ``sliding'' behavior. If the agent stops in an ineligible location, a penalty of $-0.01$ is added. 



\subsection{Model Overview}

We equip the embodied agent with different sensory modalities, and each of them feeds into a part of the input network for the \ac{rl} model. Below we introduce these components of the architecture.

\paragraph*{Visual network}


The backbone of the visual network is ResNet-18~\cite{He2016resnet} pre-trained on ImageNet. It takes $224\times224$ RGB and depth images as inputs. The weights of all layers in the network except the last fully connected layer are frozen during training. 

\begin{figure}[t!]
    \centering
    \includegraphics[width=\linewidth]{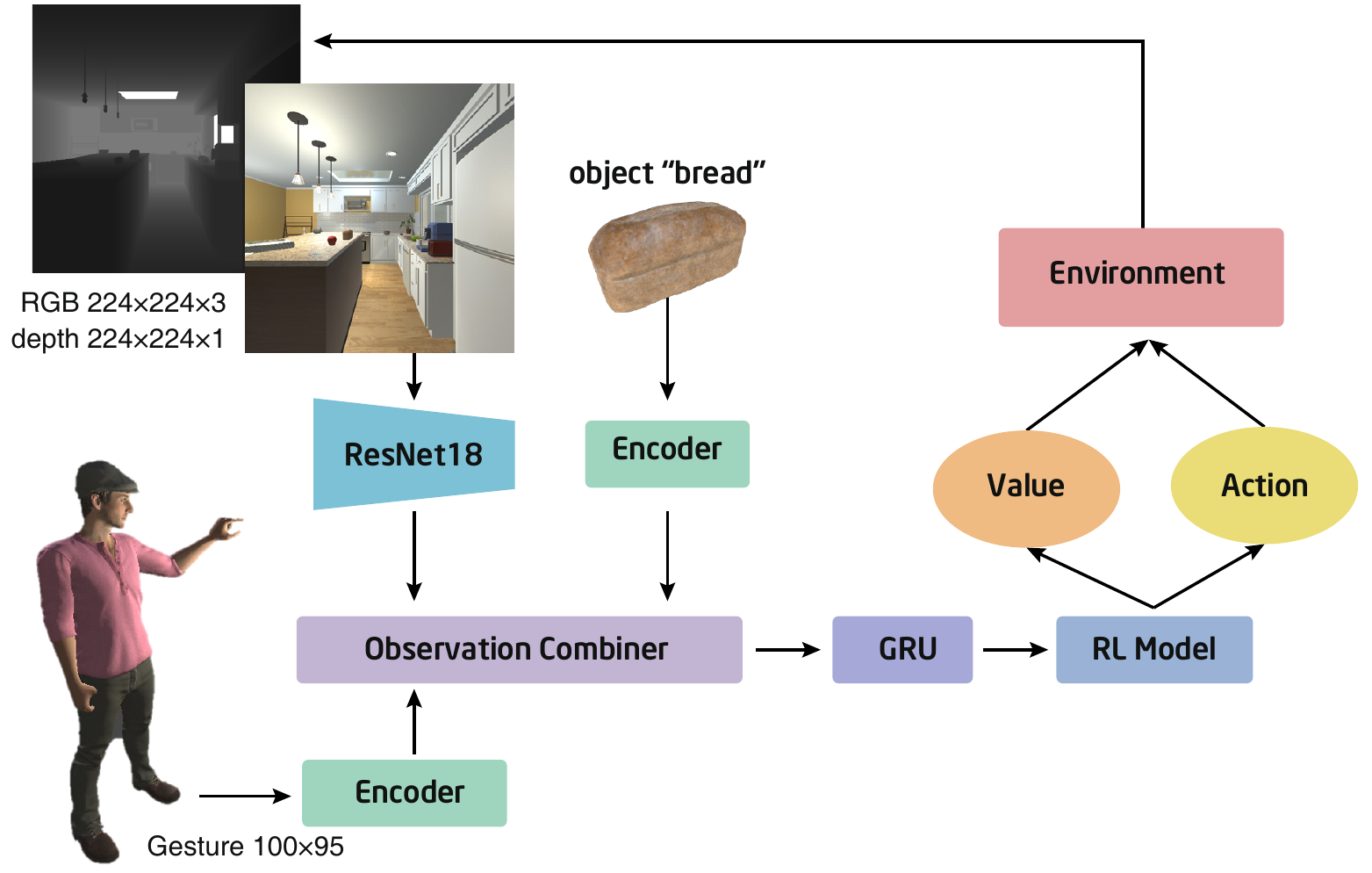}
    \caption{\textbf{The model overview.} Our model fuses perceptions from different sensory modalities, and the actor-critic model samples an action in each step according to the updated policy and send it back to the environment.}
    \label{learning_model}
\end{figure}

\paragraph*{Gesture network}


The raw input of the gesture network is a sequence with 100 time steps and 95 features. Each feature represents the muscle value from the Unity humanoid model, which can be considered as the coordinates for tracked body and hand joints. The gesture input is flattened and encoded into a vector. In addition, we provide the target object category from a selected set and pass it to an embedding layer. This is equivalent to speech or text instructions from a user in prior work on interactive embodied agent learning. Since our focus in this paper is gesture, we simplify this part of the input as a categorical variable (\eg, a single word in a fixed vocabulary). Note that this vector alone does not specify the target object location: There can be multiple instances of the same object category in the scene, and the agent needs to infer which instance the human player is referring to. This vector is concatenated with encoded gesture inputs and visual features by an observation combiner. There is a memory unit using \ac{gru}~\cite{cho2014learning} after this combiner. \cref{learning_model} illustrates the entire architecture.

\section{Evaluation}

\sisetup{detect-weight=true, detect-mode=true}
\begin{table*}[ht]
\small
    \begin{center}
    \begin{tabular}{ccS[table-format=2.1]S[table-format=2.1]S[table-format=2.1]S[table-format=2.1]S[table-format=2.1]S[table-format=2.1]}
    \toprule
    \multirow{2}[4]{*}{Scene Types} & \multirow{2}[4]{*}{Methods} & \multicolumn{3}{c}{Success Rate (\%)} & \multicolumn{3}{c}{Success weighted by Path Length (\%)} \\
    \cmidrule(lr){3-5} \cmidrule(lr){6-8}          &       & \multicolumn{1}{>{\centering}m{14mm}}{Train} & \multicolumn{1}{>{\centering}m{14mm}}{Validation} & \multicolumn{1}{>{\centering}m{14mm}}{Test} & \multicolumn{1}{>{\centering}m{14mm}}{Train} & \multicolumn{1}{>{\centering}m{14mm}}{Validation} & \multicolumn{1}{>{\centering\arraybackslash}m{14mm}}{Test} \\
    \midrule
    \multirow{3}[2]{*}{Kitchen} & \multicolumn{1}{l}{Baseline} & 12.3  & 10.2  & 11.5  & 7.6   & 7.1   & 7.9 \\
      & \multicolumn{1}{l}{Referencing} & 21.1  & 18.7  & 19.2  & 13.3  & 11.6  & 11.2 \\
      & \multicolumn{1}{l}{Intervention} & \textbf{44.9} & \textbf{31.5} & \textbf{40.3} & \textbf{27.0} & \textbf{20.0} & \textbf{24.0} \\
\midrule
\multirow{3}[2]{*}{Living Room} & \multicolumn{1}{l}{Baseline} & 6.3   & 3.6   & 3.5   & 4.1   & 2.3   & 2.1 \\
      & \multicolumn{1}{l}{Referencing} & 4.9   & 3.2   & 2.7   & 3.2   & 1.7   & 1.6 \\
      & \multicolumn{1}{l}{Intervention} & \textbf{13.0} & \bfseries 9.0 & \bfseries 9.5 & \bfseries 7.8 & \bfseries 5.3 & \bfseries 5.4 \\
\midrule
\multirow{3}[2]{*}{Bedroom} & \multicolumn{1}{l}{Baseline} & 15.2  & 9.1   & 8.7   & 9.1   & 5.3   & 5.4 \\
      & \multicolumn{1}{l}{Referencing} & \textbf{43.5}  & 10.7  & 15.4  & \textbf{28.3}  & 6.6   & 10.5 \\
      & \multicolumn{1}{l}{Intervention} & 42.4 & \textbf{22.4} & \textbf{20.4} & 27.5 & \textbf{13.8} & \textbf{11.9} \\
\midrule
\multirow{3}[2]{*}{Bathroom} & \multicolumn{1}{l}{Baseline} & 16.3  & 15.5  & 11.9  & 11.4  & 9.1   & 8.8 \\
      & \multicolumn{1}{l}{Referencing} & 33.0  & 19.4  & 19.9  & 20.5  & 11.1  & 11.7 \\
      & \multicolumn{1}{l}{Intervention} & \textbf{40.5} & \textbf{32.2} & \textbf{35.0} & \textbf{29.1} & \textbf{21.0} & \textbf{23.0} \\
\midrule
\multirow{4}[1]{*}{Average} & \multicolumn{1}{l}{Baseline} & 12.5  & 8.1   & 9.9   & 6.2   & 8.6   & 5.9 \\
      & \multicolumn{1}{l}{Referencing} & 25.6  & 16.3  & 13.0  & 7.8   & 14.3  & 8.8 \\
      & \multicolumn{1}{l}{Intervention} & \textbf{35.2} & \textbf{22.9} & \textbf{23.8} & \textbf{15.0} & \textbf{26.3} & \textbf{16.1} \\
& Scene Prior~\cite{yang2018visual}$^*$  &       &       & 13.4  &       &       & 6.7 \\
    \bottomrule
\end{tabular}%

    \end{center}
    \caption{\textbf{Evaluation results for train/validation/test split.} \ac{sr} and \ac{spl} at the first stop are reported in this table. We compare models trained with referencing gestures and intervention gestures against a baseline model. 
    $^*$ Reported from~\cite{yang2018visual}. This method uses additional scene prior knowledge but not gestures. 
    }
    \label{table1}
\end{table*}

\begin{table*}[t]
\small
    \begin{center}
\begin{tabular}{clS[table-format=2.1]S[table-format=2.1]S[table-format=2.1]S[table-format=2.1]S[table-format=2.1]S[table-format=2.1]S[table-format=2.1]S[table-format=2.1]}
\toprule
\multirow{2}[1]{*}{Scene Types} & \multicolumn{1}{c}{\multirow{2}[1]{*}{Methods}} & \multicolumn{4}{c}{Success Rate (\%)} & \multicolumn{4}{c}{Success weighted by Path Length (\%)} \\
\cmidrule(lr){3-6} \cmidrule(lr){7-10}       &       & \multicolumn{1}{>{\centering}m{11.5mm}}{1 Stop} & \multicolumn{1}{>{\centering}m{11.5mm}}{2 Stop} & \multicolumn{1}{>{\centering}m{11.5mm}}{3 Stop} & \multicolumn{1}{>{\centering}m{11.5mm}}{$\infty$} & \multicolumn{1}{>{\centering}m{11.5mm}}{1 Stop} & \multicolumn{1}{>{\centering}m{11.5mm}}{2 Stop} & \multicolumn{1}{>{\centering}m{11.5mm}}{3 Stop} & \multicolumn{1}{>{\centering}m{11.5mm}}{$\infty$} \\
\midrule
    \multirow{3}[2]{*}{Kitchen} & Baseline & 11.5  & 18.0  & 23.1  & 49.3  & 7.9   & 12.1  & 14.6  & 25.1 \\
          & Referencing & 19.2  & 26.3  & 29.7  & 47.9  & 11.2  & 15.0  & 16.6  & 24.1 \\
          & Intervention & \textbf{40.3} & \textbf{55.1} & \textbf{62.8} & \textbf{89.0} & \textbf{24.0} & \textbf{32.7} & \textbf{37.2} & \textbf{52.8} \\
    \midrule
    \multirow{3}[2]{*}{Living Room} & Baseline & 3.5   & 6.4   & 9.5   & 23.3  & 2.1   & 3.7   & 5.1   & 9.8 \\
          & Referencing & 2.7   & 3.9   & 4.7   & 7.8   & 1.6   & 2.4   & 2.9   & 4.7 \\
          & Intervention & \bfseries 9.5 & \textbf{16.9} & \textbf{21.7} & \textbf{58.0} & \bfseries 5.4 & \bfseries 9.8 & \textbf{12.6} & \textbf{30.8} \\
    \midrule
    \multirow{3}[2]{*}{Bedroom} & Baseline & 8.7   & 15.3  & 18.4  & 31.0  & 5.4   & 9.5   & 11.2  & 17.0 \\
          & Referencing & 15.4  & 18.7  & 20.2  & 31.9  & 10.5  & 12.4  & 13.3  & 19.2 \\
          & Intervention & \textbf{20.4} & \textbf{27.9} & \textbf{33.5} & \textbf{51.2} & \textbf{11.9} & \textbf{16.3} & \textbf{19.5} & \textbf{29.7} \\
    \midrule
    \multirow{3}[2]{*}{Bathroom} & Baseline & 11.9  & 18.9  & 23.7  & 57.8  & 8.8   & 13.8  & 16.9  & 33.1 \\
          & Referencing & 19.9  & 30.5  & 35.9  & 64.3  & 11.7  & 18.1  & 21.3  & 32.4 \\
          & Intervention & \textbf{35.0} & \textbf{44.6} & \textbf{51.7} & \textbf{76.5} & \textbf{23.0} & \textbf{29.6} & \textbf{33.9} & \textbf{48.3} \\
    \midrule
    \multirow{3}[2]{*}{Average} & Baseline & 8.9   & 14.7  & 18.7  & 40.4  & 6.1   & 9.8   & 12.0  & 21.3 \\
          & Referencing & 14.3  & 19.9  & 22.6  & 38.0  & 8.8   & 12.0  & 13.5  & 20.1 \\
          & Intervention & \textbf{26.3} & \textbf{36.1} & \textbf{42.4} & \textbf{68.7} & \textbf{16.1} & \textbf{22.1} & \textbf{25.8} & \textbf{40.4} \\
    \bottomrule
\end{tabular}%
    \end{center}
    \caption{\textbf{Evaluation results for test scenes with different number of allowed stops ($\infty$ denotes infinte allowed stops).} \ac{sr} and \ac{spl} are presented. We compare models trained with referencing gestures and intervention gestures against a baseline model.}
    \label{table2}
\end{table*}

We evaluate our methods in \ac{gesthor} environment. AI2-THOR provides 120 scenes covering four different room types: kitchen, living room, bedroom, and bathroom. Each room has its own unique appearance and arrangements. We randomly split 30 scenes for each scene type into 20 training rooms, 5 validation rooms, and 5 testing rooms.


There are 38 object categories available for all scenes. Since there is almost no overlap of objects for different scene types, we train and evaluate separately for each scene type. We evaluate each scene for 250 episodes and report the average results for each scene type.

\paragraph*{Evaluation Metrics}

We use 2 metrics to evaluate different methods:

\begin{itemize}[leftmargin=*,noitemsep,nolistsep,topsep=0pt]
    \item \ac{sr}: for the $i$-th episode, the success can be marked by a binary indicator $S_{i}$. The success rate is the ratio of successful episodes over completed episodes N:
    \begin{equation}
        SR = \frac{1}{N}\sum_{i=1}^{N}S_{i}.
    \end{equation}
    
    \item \ac{spl}: this metric is proposed by Anderson \etal~\cite{anderson2018evaluation}. It measures the efficacy of the navigation. \ac{spl} is calculated as follows:
    \begin{equation} \label{SMS}
        SPL = \frac{1}{N}\sum_{i=1}^{N}S_{i}\left( \frac{l_i}{\max(p_{i},l_{i})} \right),
    \end{equation}
    where $l_{i}$ is the shortest path distance from the agent's starting position to the goal in episode $i$, and $p_{i}$ is the actually path length taken by the agent.
\end{itemize}



We have three methods to evaluate the agent performance: (1) Baseline: the agent only has the visual (\ie, RGB and depth images) and object category information. (2) Referencing Gesture: in addition to (1), the agent receives referencing gesture inputs. (3) Intervention Gesture: in addition to (1), the agent receives rejective gesture inputs when the forward direction forms an angle larger than 90 degrees between the agent and the target. 

In our comparative setting, the baseline model does not use any gestures. While one may expect that it should always underperform, this is only true if the agent has learned and inferred the semantics of human gestures and incorporated the signals during navigation, which is the focus of our evaluation. Again, this is not trivial because we do not pre-define the meaning of any gestures. Similarly, the intervention gestures is a strong directive feedback from the human user, but we evaluate how well the agent can infer its meaning and adopt it in navigation. 

\paragraph*{Navigation Performance}

\cref{table1} show the performance of different methods when evaluated at the first stop, and \cref{table2} show the performance at test scenes evaluated at a different number of stops. From the both results, we confirm that adding gestures can significantly improve the navigation success rate as well as the efficiency over the baseline model. \cref{table1} puts a hard constraint on the number of stops to 1 to match the state-of-art benchmarks~\cite{anderson2018evaluation}. Of note, models trained with intervention gestures outperform models trained with referencing gestures, both in \ac{sr} and \ac{spl}, demonstrating that intervention gesture is a more effective kind of gesture to communicate with the agent. 
\cref{table2} reports results on test scenes with a different number of allowed stops. We should see that both \ac{sr} and \ac{spl} increase with the number of allowed stops, and the improvement of \ac{sr} and \ac{spl} with gestures is more evident in a lower number of allowed stops.

\paragraph*{Qualitative Results}

\begin{figure}[t!]
    \centering
    \includegraphics[width=\linewidth]{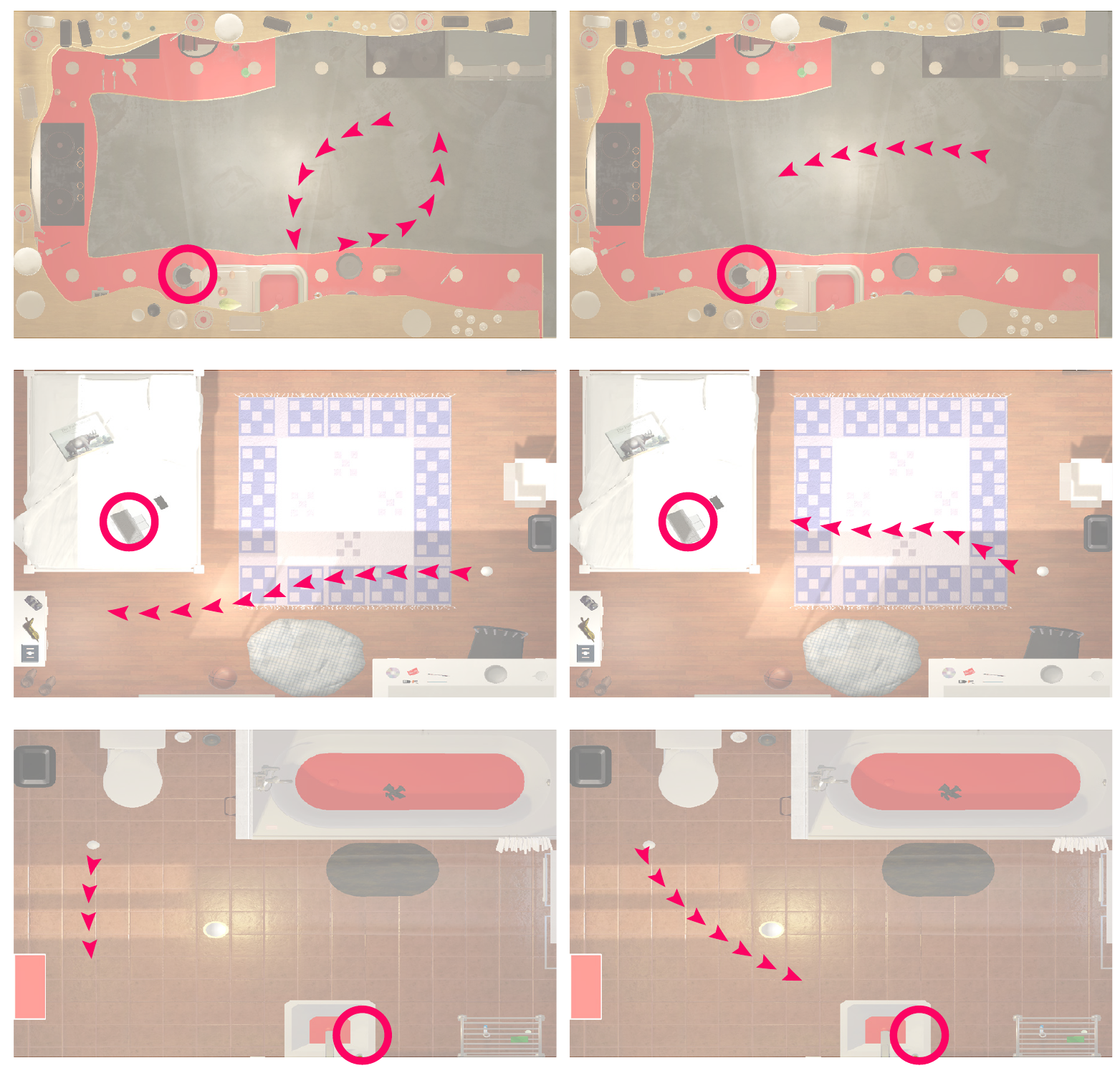}
    \caption{\textbf{Qualitative results with visualizations of trajectories for baseline (left) and referencing gesture (right) models.} Our agent can efficiently navigate to the target with the help of gestures.}
    \label{fig:referencing_gesture}
\end{figure}

\begin{figure}[t!]
    \centering
    \includegraphics[width=\linewidth]{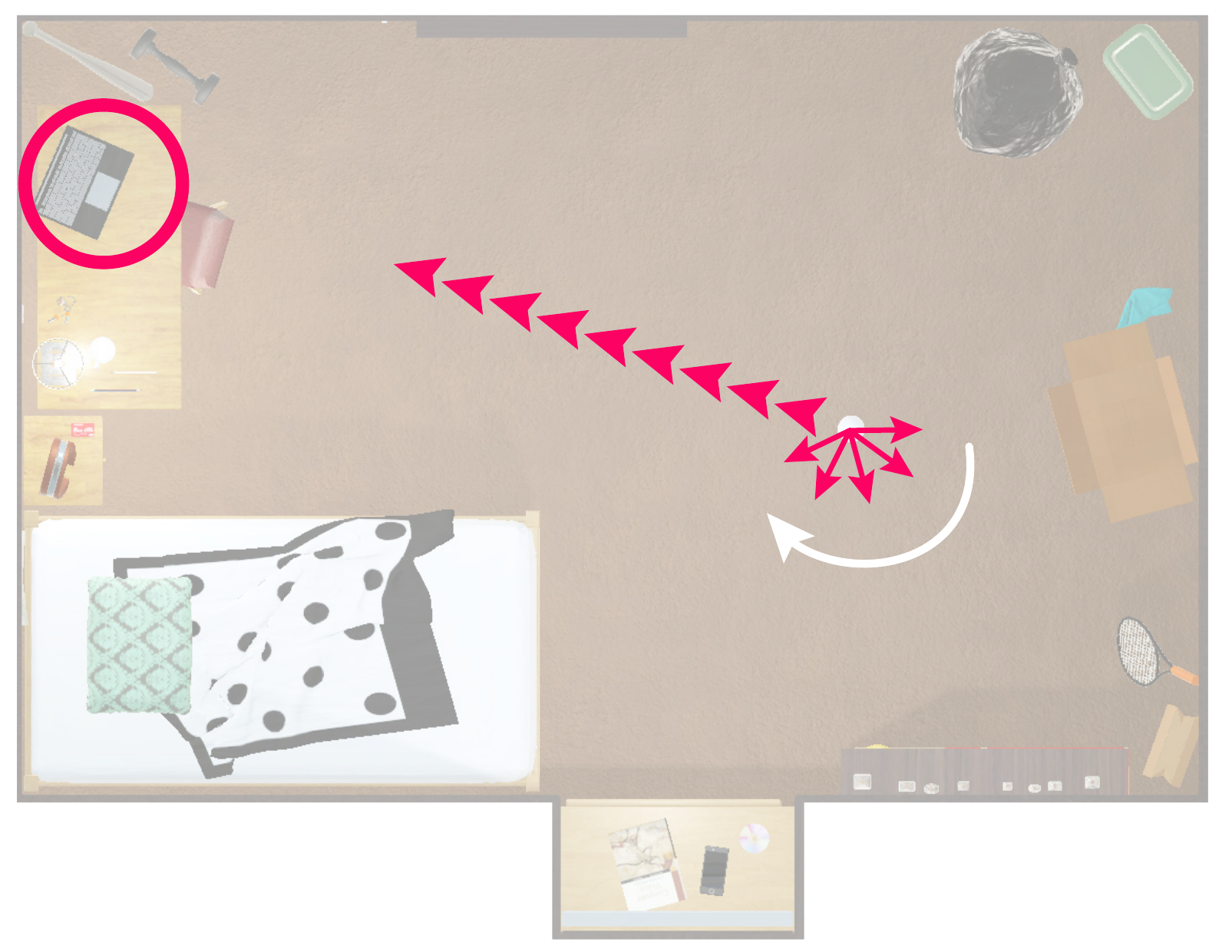}
    \caption{\textbf{Qualitative results for the intervention gesture model.} When the agent is back to the target, it receives interventional gestures and could first rotate until it faces the target before making any forward movements, thus increasing the navigation success rate and efficiency.}
    \label{fig:intervention_gesture}
\end{figure}

To visualize the effectiveness of our methods, we show some qualitative results in \cref{fig:referencing_gesture,fig:intervention_gesture}. \cref{fig:referencing_gesture} compares our referencing gesture model against the baseline model with visualized trajectories in different scenes and targets. It could be observed that in all scenes, our referencing gesture model enables the agent to navigate to the target more intelligently, while the baseline model often struggles to find the target and stop or takes a longer path to find the target. \cref{fig:intervention_gesture} demonstrates how our intervention gesture model works to improve the navigation significantly. In this example, the agent rotates at the place where it faces back to the target and is instructed with interventions gestures until the target is in its field of view before making any movements. This indicates that our agent is able to understand and react to the intervention gestures, resulting in much better navigation performance.

\section{Conclusion}

In this paper, we propose a new framework for embodied visual navigation where human users can give instructions to the autonomous agent using gestures. Such agents and gesture based interface will be very useful for collaborative robots or virtual agents. We have built a VR-based interactive learning environment, \ac{gesthor}, based on AI2-THOR and designed an end-to-end deep reinforcement learning model for the navigation task. Our experiments show that the agent is able to interpret human instructions with gestures and improve its visual navigation. We also conclude that interactive activities during agent task execution can improve performance. While the main setting and experimental design of our study have been used in prior works, to the best of our knowledge, our paper is the first incorporating human gestures for embodied agent learning and showing the agent can learn the semantic of gestures without supervision. We will make publicly available our simulation environment and the recorded gesture dataset for any future research for Human-AI interaction via gestures. The future directions include adding more objects, tasks, gestures, and multiple agents in the scene, \eg, navigating to an object and bring it back by showing gestures in our framework and also allowing agents to make gestures to the human player such that both parties can communicate with gestures, which will also help humans to utilize even more diverse gestures to communicate with agents. 


\setstretch{0.94}
\small
\balance
\bibliographystyle{ieeetr}
\bibliography{reference}
\end{document}